# Improved Information Gain Estimates for Decision Tree Induction

**Sebastian Nowozin**  Sebastian.Nowozin@microsoft.com
Microsoft Research Cambridge, United Kingdom

## Abstract

Ensembles of classification and regression trees remain popular machine learning methods because they define flexible non-parametric models that predict well and are computationally efficient both during training and testing. During induction of decision trees one aims to find predicates that are maximally informative about the prediction target. To select good predicates most approaches estimate an information-theoretic scoring function, the *information gain*, both for classification and regression problems. We point out that the common estimation procedures are biased and show that by replacing them with improved estimators of the discrete and the differential entropy we can obtain better decision trees. In effect our modifications yield improved predictive performance and are simple to implement in any decision tree code.

## 1. Introduction

Decision trees are a classic method of inductive inference that is still very popular (Breiman et al., 1984; Hunt et al., 1966). They are not only easy to implement and use for classification and regression tasks, but also offer good predictive performance, computational efficiency, and flexibility. While often heuristically motivated, they can also deal with mixed discrete/continuous features and missing values. As early as in the late 1970's ID3 introduced the use of information theory for the induction of decision trees from data, described in detail in (Quinlan, 1986). Yet, optimal induction of decision trees according to some global objective is fundamentally hard (Hyafil & Rivest, 1976), and to this day most implementations use randomized greedy algorithms for growing a decision tree (Criminisi et al., 2012). A recent advance has been the use of ensembles of randomized decision trees (Amit & Geman, 1994), such as *random forests* (Breiman, 2001), improving the predictive performance by averaging predictions of multiple trees.

Reflecting on this rich history, we ask an almost old-fashioned question: Can we improve the quality of learned decision trees by better estimation of information-theoretic quantities? To answer this, we take a look at how decision trees are typically grown and how information theory is used in the process.

### 1.1. Decision Tree Induction

Decision trees are most often induced greedily, in the following manner. Given a data set $\{(x_i, y_i)\}_{i=1}^{N}$, we start with an empty tree with just a root node. We then sample a number of split function candidates from a fixed distribution. Each split function partitions the training set into a left and right subset by some test on each $x_i$. These subsets are scored so that a high score is assigned to splits that aid in predicting the output $y$ well, i.e. those that reduce the average uncertainty about the predictive target as estimated by the training set. The highest scoring split is selected and the training set is partitioned accordingly into the two child nodes, growing the tree by making the node the parent of the two newly created child nodes. This procedure is applied recursively until some stopping conditions such as a maximum tree depth or minimum sample size are reached, see e.g. (Breiman et al., 1984).

Different split scoring functions exist and have been used in the past (Breiman et al., 1984). A popular split scoring function, the so called *information gain* is derived from information-theoretic considerations. The information gain is just the *mutual information* between the local node decision (left or right) and the predictive output.

To motivate the information gain score, let $b \in \mathcal{B}$ be a random variable denoting the decision whether a sample is assigned to the left (L) or right (R) branch by the split function. Hence $\mathcal{B} = \{L, R\}$, and there is a joint distribution $p(y, b)$. Once we fix a sample $x$

Improved Information Gain Estimates for Decision Tree Induction**Algorithm 1** Information gain split selection
1: **Input:** Sample set $Z = \{(x_i, y_i)\}_{i=1}^{n_\ell}$, split proposal distribution $p_s$, number of tests $T \geq 1$, entropy estimator $\hat{H}$
2: **Output:** Split function $s^* : \mathcal{X} \to \{L, R\}$
3: **Algorithm:**
4: $g^* \leftarrow 0$ {Initialize best infogain}
5: **for** $t = 1, \ldots, T$ **do**
6: $\quad$ Sample binary split $s \sim p_s$
7: $\quad$ Partition $Z$ into $(Z_L, Z_R)$ using $s$
8: $\quad$ $g \leftarrow -\frac{|Z_L|}{n_\ell}\hat{H}(Y_L) - \frac{|Z_R|}{n_\ell}\hat{H}(Y_R)$ {Estimate}
9: $\quad$ **if** $g > g^*$ **then**
10: $\quad\quad$ $(g^*, s^*) \leftarrow (g, s)$ {New best infogain}
11: $\quad$ **end if**
12: **end for**

the variable $b$ is deterministic. We now measure the *mutual information* (Cover & Thomas, 2006) between the variable $b$ and $y$. From the definition of mutual information and by some basic manipulation we have

$$\begin{aligned} I(y, b) &= D_{KL}(p(y, b) \parallel p(y)p(b)) \quad &(1)\\ &= \int_\mathcal{B} \int_\mathcal{Y} p(y|b)p(b) \log p(y|b) \, \mathrm{d}y \, \mathrm{d}b \\ &\quad - \int_\mathcal{B} \int_\mathcal{Y} p(y, b) \log p(y) \, \mathrm{d}y \, \mathrm{d}b \quad &(2)\\ &= H_y - \sum_{b \in \{L, R\}} p(b) \, H_{y|b}. \quad &(3) \end{aligned}$$

Because the entropies $H_y$, $H_{y|b}$, and the split marginal $p(b)$ are unknown, they are typically estimated using the training sample. Depending on $\mathcal{Y}$, the measure of integration $\mathrm{d}y$ is either discrete (classification) or the Lebesgue measure (regression). Therefore the entropies are either discrete entropies or *differential entropies* (Cover & Thomas, 2006). The information gain (3) is a popular criterion used to determine the quality of a split (Criminisi et al., 2012), and has been used for classification, regression, and density estimation. It is generally accepted as the criterion of choice except for very unbalanced classification tasks, or when the noise on the prediction targets $y$ is large.

When we use the information gain criterion for recursive tree growing, we execute Algorithm 1 at each node in the decision tree, testing $T$ candidate splits sequentially and keeping the one that achieves the highest estimated information gain. In the algorithm $Y_L$ is the subset of training set predictions that are sorted into the left branch, i.e. $Y_L = \{y_i : s(x_i) = L\}$. Likewise $Y_R$ is the remaining set of predictions. After the best split is selected the left and right child nodes are again subjected to the algorithm.

In this work we assume that the information gain (3) is a good criterion by which to select splits. Given this assumption, we address the problem of how to *estimate* the information gain from a finite sample. As we will see, the currently used estimators can be improved, yielding more accurate estimates of the information gain and in turn better decision trees.

### 1.2. Related work

The importance of how a split proposal is scored has been analyzed in a series of papers. (Mingers, 1989) provides an initial analysis and state that the "choice of measure affects the size of a tree but not its accuracy". At that time this was considered a controversial claim and indeed (Buntine & Niblett, 1992) and (Liu & White, 1994) demonstrated that the statistical analysis of Mingers was flawed and the different methods of scoring a split do have a substantial influence on the generalization error of the learned decision tree.

For binary classification tasks the choice of split scoring functions has attracted special attention, as in this case the split scoring function can be related to statistical *proper scoring rules* (Buja et al., 2005). In addition, specialized criteria aiding the interpretability of the resulting tree have been developed (Buja & Lee, 2001). In our work we consider general multiclass classification and multivariate regression tasks where the goal is good predictive performance.

## 2. Multiclass Classification

For multiclass classification, we predict into a finite set of classes $\mathcal{Y} = \{1, 2, \ldots, K\}$. The most commonly used entropy estimator is derived from *empirical class probabilities*. For each class $k$, we count the number of occurences of that class as $h_k = \sum_{y \in Y} I(y = k)$. The sum of all counts is $n = |Y| = \sum_k h_k$.

### 2.1. Naive Entropy Estimate

To estimate the entropy, we use the empirical class probabilities $\hat{p}_k(Y) = h_k/n$, and the estimate

$$\begin{aligned} \hat{H}_N(Y) &= -\sum_{k=1}^K \hat{p}_k(Y) \log \hat{p}_k(Y) \quad &(4)\\ &= \log n - \frac{1}{n} \sum_{k=1}^K h_k \log h_k, \quad &(5) \end{aligned}$$

where in case a term is $0 \log 0$ it is taken to be zero.

The entropy estimator $\hat{H}_N$ is an instance of a *plug-in* estimator, where a function is evaluated on an *estimated* probability distribution. In the case of the dis-

# Improved Information Gain Estimates for Decision Tree Induction

crete entropy this is consistent, that is, in the large sample limit $n \to \infty$ it converges to the true entropy (Antos & Kontoyiannis, 2001). Although (4) is popularly used (see e.g. (Criminisi et al., 2012) and references therein), it is a bad estimator when used in (3). In particular (4) is *biased* and universally underestimates the true entropy. In fact, the bias is known (Schürmann, 2004), and we have for $n = |Y|$ samples that the bias depends on the true but unknown per-class probabilities $p_k$ as

$$H(Y) - \mathbb{E}[\hat{H}_N(Y)] = \frac{K-1}{2n} - \frac{1}{12n^2}\left(1 - \sum_{k=1}^{K} \frac{1}{p_k}\right) + O(n^{-3}).$$

Unfortunately, the second and higher-order terms of this bias depend on the true unknown probabilities. The first term, $\frac{K-1}{2n}$, however, can be evaluated and is known as the *Miller correction* to (4), see (Miller, 1955). The correction is useful for entropy estimation, but has no effect in (3). To see this, let us define the Miller-adjusted estimate, $\hat{H}_M(Y) = \hat{H}_N(Y) + \frac{K-1}{2|Y|}$, and use it to evaluate the sum in (3), as

$$\begin{aligned} & -\frac{|Z_L|}{n_\ell}\hat{H}_M(Y_L) - \frac{|Z_R|}{n_\ell}\hat{H}_M(Y_R) \\ =\ & -\left(\frac{|Z_L|}{n_\ell}\hat{H}_N(Y_L) + \frac{|Z_R|}{n_\ell}\hat{H}_N(Y_R)\right) + C, \end{aligned}$$

where the constant $C = (K-1)/n_\ell$ is independent of the split and thus it influences all split scores in the same way in (3). Therefore, although the entropy estimates are improved in absolute accuracy, the Miller correction has no effect on which split is selected.

Can we at all hope to achieve a better split from using improved entropy estimators? Estimating the discrete entropy is challenging (Paninski, 2003; Antos & Kontoyiannis, 2001), but superior corrections than the one of Miller have been proposed recently. We cannot give an overview of the many proposed estimators here, but recommend the work (Schürmann, 2004) for an overview and quantitative comparison. For our purposes, we select the Grassberger entropy estimate.

## 2.2. Grassberger Entropy Estimate

In two papers (Grassberger, 1988; 2003), Peter Grassberger derived a family of discrete entropy estimators. We use the refined estimator from 2003, given as

$$\hat{H}_G(h) = \log n - \frac{1}{n}\sum_{k=1}^{K} h_k G(h_k), \qquad (6)$$

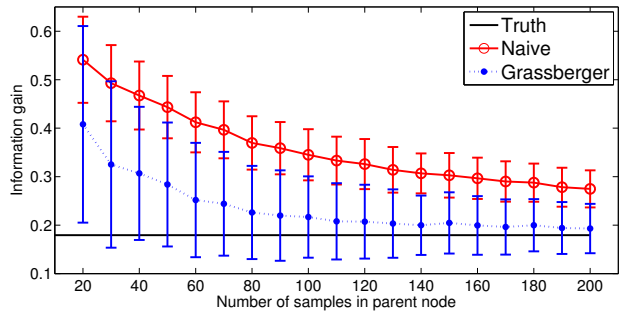

Figure 1. Illustration of information gain estimation. For a 40-class problem we produce finite-sample histograms, varying the total number of samples in the set. The set is partitioned into a left and right subset according to a fixed multinomial distribution over classes, yielding a positive true information gain (which we can compute in this example because we know the multinomial distribution). Shown are the mean estimates and one unit standard deviation over 500 replicates. The Grassberger entropy estimator $\hat{H}_G$ produces more accurate estimates than the naive plugin entropy estimator.

where $n = |Y|$, and the summation is over all $h_k > 0$ in the histogram. The function $G(h)$ is given as[1]

$$G(h_k) = \psi(h_k) + \frac{1}{2}(-1)^{h_k}\left(\psi(\frac{h_k+1}{2}) - \psi(\frac{h_k}{2})\right).$$

In the above formula, $\psi$ is the *digamma* function. For large $h_k$ the function $G(h_k)$ behaves like a logarithm, and hence in the limit $n \to \infty$ we see that (6) becomes identical to (5). For small sample sizes $G(h_k)$ differs from $\log h_k$ and the resulting estimate $\hat{H}_G$ is more accurate (Schürmann, 2004).

To train classification trees using $\hat{H}_G$ we simply use it instead of $\hat{H}_N$ in Algorithm 1. The improvement in information gain estimation is dramatic, as shown in Figure 1. It is now fair to ask whether this improvement in information gain estimation also leads to better generalization performance in the final classification tree ensemble. We will address this experimentally in the next section, but generally we expect the improvement to be largest when many classes are used (say, $K \geq 10$), because in this case the empirical class frequencies cannot be estimated reliably.

## 2.3. Experiments

For the experimental protocol we follow the recommendations in (Demšar, 2006). We intentionally do not compare against other classification methods, both because the competitive performance of randomized

---

[1]This is the closed-form solution to equation (30) in (Grassberger, 2003), where $G(h_k)$ was given only as $G(h_k) = \psi(h_k) + (-1)^{h_k}\int_0^1 \frac{x^{h_k-1}}{x+1}dx$.

# Improved Information Gain Estimates for Decision Tree Induction

decision trees has been established before and because we want to assess only the contribution of improved information gain estimates. Overall, we use a standard setup for training our classification trees; no pruning is used and there are no missing features.

As features we use simple single-dimension thresholding tests of the form $s(x, a, b) = (x_a \leq b)$, where $x_a$ is the $a$'th element in vector $x$, and $b$ is a threshold. The proposal distribution at a node is to select $a$ uniformly at random, but $b$ is sampled uniformly from the samples reaching that node. At the leaf nodes we store a single class label, the majority class over all remaining samples at that node. Ties are broken at random. The ensemble of trees is trained using simple replication of the training set. To make predictions at test time, we simply take the majority decision over the individual predictions made by all trees in the ensemble.

**Setup.** We use 30 data sets from a variety of sources. We use all multiclass ($K \geq 3$) data sets from UCI, as well as four text classification data sets (`reuters`, `tdt2`, `news20`, `rcv1`) and web data sets.[2] For data sets that come with a fixed `train/val/test` split we have used this split. For data sets that only come with a `train/test` split we further split the original `train` set 50/50 into a new `train/val` set. For data sets that do not provide any split we partition the data into the three subsets according to the proportions 25/25/50. We perform model selection entirely on the `val` set, selecting the minimum number of samples required to continue splitting in $\{1, 5, 10\}$, but fixing the number of trees to eight and the number of feature tests to 256. The best parameter on the validation set is taken and one model is trained on `train+val`, then evaluated on `test`. We repeat the procedure five times and report averaged multiclass accuracies on the `test` set.

**Results.** Table 1 shows the average multiclass accuracies achieved on the `test` set. Generally the performance appears to be on the same level, with some notable exceptions (`tdt2`, `news20`, `rcv1`, and `sector`) where the Grassberger entropy estimator obtains higher accuracies. A Wilcoxon signed-rank test rejects the null hypothesis of equal performance at a p-value of 0.0267. Overall, the Grassberger estimate yields higher accuracies on 18 data sets, compared to eight data sets for the naive estimate, with four ties. The runtime of the two methods showed no difference and the the entire experiment ran on a single eight core PC in less than 10 hours.

---

[2]From http://www.zjucadcg.cn/dengcai/Data/TextData.html and http://www.cs.umd.edu/~sen/lbc-proj/LBC.html

Table 1. Multi-class classification accuracies achieved on different data sets. We report the average test set accuracy as well as one unit standard deviation over five replicates. Comparing the mean results over all data sets, a Wilcoxon signed-rank test (Demšar, 2006) rejects the null-hypothesis that the classifiers perform equally well when using $\hat{H}_G$ and when using $\hat{H}$, at a p-value of 0.0267. For each data set, we print the highest average performance in boldface.

| Data set | Classes | Using $\hat{H}$ | Using $\hat{H}_G$ |
|---|---|---|---|
| WEBKB | 5 | **77.7±2.0** | 76.7±3.9 |
| CORA | 7 | 67.1±0.9 | **67.9±0.9** |
| CITESEER | 6 | 66.3±0.8 | **66.4±0.9** |
| REUTERS | 58 | 83.3±0.6 | **84.7±0.5** |
| TDT2 | 96 | 88.9±0.7 | **90.8±0.6** |
| NEWS20 | 20 | 64.8±0.4 | **68.3±0.7** |
| RCV1 | 53 | 70.4±0.3 | **72.1±0.4** |
| FACES-OLIVETTI | 40 | **85.9±3.3** | 85.5±3.2 |
| MNIST | 9 | 94.5±0.08 | 94.5±0.1 |
| USPS | 10 | 91.7±0.4 | 91.7±0.4 |
| CONNECT4 | 3 | 76.2±0.4 | **76.7±0.7** |
| COVTYPE | 7 | **96.0±0.04** | 95.9±0.04 |
| DNA | 3 | 93.0±0.4 | **93.4±0.4** |
| GLASS | 6 | **66.1±4.7** | 65.1±4.9 |
| IRIS | 3 | 93.3±1.8 | **93.5±2.6** |
| LETTER | 26 | **88.3±0.3** | 88.2±0.4 |
| OILFLOW | 3 | 99.0±0.3 | **99.2±0.3** |
| PENDIGITS | 9 | 95.5±0.4 | **95.7±0.6** |
| POKER | 9 | **56.4±0.1** | 56.2±0.1 |
| PROTEIN | 3 | **61.4±0.4** | 61.1±0.6 |
| SATIMAGE | 6 | 89.1±0.6 | **89.2±0.4** |
| SECTOR | 105 | 49.1±1.1 | **54.9±1.3** |
| SEISMIC | 3 | 75.2±0.2 | 75.2±0.1 |
| SHUTTLE | 7 | 100.0±0.0 | 100.0±0.1 |
| SOY | 3 | 90.1±3.0 | **91.3±1.2** |
| SVMGUIDE2 | 3 | 71.6±2.7 | **72.5±2.0** |
| SVMGUIDE4 | 3 | **78.6±1.6** | 78.2±1.5 |
| VEHICLE | 4 | 70.6±2.0 | **72.2±1.4** |
| VOWEL | 10 | 44.5±1.6 | **44.8±2.2** |
| WINE | 3 | 90.0±3.7 | **91.2±3.8** |
| WINS (4 TIES) | | 8 | **18** |

**Discussion.** The difference between the two estimators is small but statistically significant and we can conclude that improved entropy estimation yields improved classification trees. All other things have remained unchanged and therefore the difference is directly attributable to the entropy estimation. The change is most pronounced on data sets with a large number of classes, confirming the superiority of the Grassberger estimate in these situations. Implementation-wise this improvement comes for free: all that is required is to replace the naive entropy estimate with the Grassberger estimate.

## 3. Multivariate Regression

For multivariate regression problems and density estimation tasks we have $\mathcal{Y} = \mathbb{R}^d$. Moreover, we require that for the true generating distribution $q(x,y)$ we have that $q(y|x)$ is a density with respect to the Lebesgue measure, that is, the CDF of $q(y|x)$ is absolutely continuous. Our goal is to learn a prediction model $p(y|x)$ by evaluating a decision tree and storing a simple density model $p_\ell(y)$ at the leaf $\ell$.

For any such distribution we consider its *differential entropy* (Cover & Thomas, 2006) as a measure of uncertainty, $H(q) = -\int_y q(y|x) \log q(y|x) \mathrm{d}y$. In general, estimating the differential entropy of an unknown continuous distribution is a more difficult problem than in the discrete case. The next subsection discusses the common approach for training regression trees before we propose a better estimator.

### 3.1. Normal approximation

One strategy is to assume the sample is a realization of a distribution within a known parametric family of distributions. By identifying a member in this family using a point estimate we can compute the entropy in closed form analytically. The most popular multivariate distribution that is amenable to such analytic treatment is the multivariate Normal distribution. For example, (5.2) in (Criminisi et al., 2012) uses the entropy of a multivariate Normal in $d$ dimensions,

$$\hat{H}_{\text{MVN-PLUGIN}}(\hat{C}(Y)) = \frac{d}{2} - \frac{d}{2}\log(2\pi) + \frac{1}{2}\log|\hat{C}(Y)| \quad (7)$$

where $\hat{C}(Y)$ is the sample covariance matrix. This is a consistent *plugin-estimator* of the entropy because $\hat{C}(Y)$ is a consistent estimator of the covariance parameter of the Normal distribution. A special case is the diagonal approximation $\hat{H}_{\text{DIAG}}(C) = \hat{H}_{\text{MVN-PLUGIN}}(C \odot I)$ where only the variances are used and $\odot$ is the elementwise matrix product, such that all covariances are set to zero. However, as for the discrete entropy, this estimator is biased: if the true distribution is Normal, then (7) underestimates its entropy.

If we are to accept the Normal approximation, we should use a better estimator of its entropy. Such an estimator exists, as (Ahmed & Gokhale, 1989) show: they derive among all unbiased estimators one that uniformly has the smallest variance. It is

$$\hat{H}_{\text{MVN-UMVUE}} = \frac{d}{2}\log(e\pi) + \frac{1}{2}\log|\sum_{y \in Y} yy^T| \quad (8)$$
$$-\frac{1}{2}\sum_{j=1}^{d} \psi\left(\frac{n+1-j}{2}\right),$$

where $\psi$ is again the *digamma* function. While (8) improves over (7), a fundamental problem with the Normal approximation is *misspecification*: we assume that the samples come from a Normal distribution, but in fact they could come from any other distribution.

### 3.2. Non-parametric Entropy Estimation

Instead of making the assumption that the distribution belongs to a parametric family of distributions, *non-parametric entropy estimates* aim to work for all distributions satisfying some general assumptions. Typically, the assumptions we make here, bounded support and absolute continuity, are sufficient conditions for these estimates to exist.

The *1-nearest neighbor estimator* of the differential entropy was proposed by (Kozachenko & Leonenko, 1987) (a detailed description in English is given in (Beirlant et al., 2001)). For each sample we define the one nearest neighbor (1NN) distance $\rho_i = \min_{j \in \{1,\dots,n\} \setminus \{i\}} \|y_j - y_i\|$. The estimate is

$$\hat{H}_{1\text{NN}} = \frac{d}{n}\sum_{i=1}^{n} \log \rho_i + \log(n-1) + \gamma + \log V_d, \quad (9)$$

where $\gamma \approx 0.5772$ is the Euler-Mascheroni constant, and $V_d = \pi^{d/2}/\Gamma(1 + \frac{d}{2})$ is the volume of the $d$-dimensional hypersphere. We can efficiently compute (9) for small number of output dimensions (say, $d \leq 10$) by the use of k-d trees (Friedman et al., 1977). Then, computing $\rho_i$ is $O(\log n)$ so that for a fixed dimension we can evaluate $\hat{H}_{1\text{NN}}$ in $O(n \log n)$ time. To speed up evaluation, we subsample the set $Y$ used in $H_{1\text{NN}}$ to keep only 256 samples, without replacement.

Other non-parametric estimates of the differential entropy have been derived from kernel density estimates (Beirlant et al., 2001), length of minimum spanning trees (Hero & Michel, 1999; Costa & Hero, 2004), and k-nearest neighbor distances (Goria et al., 2005). The latter two could potentially improve on the 1-NN estimate, but we did not examine them further. For an introduction to the field of non-parametric estimation of information-theoretic functions, see (Wang et al., 2009), and the earlier survey (Beirlant et al., 2001).

### 3.3. Experiments

For the experiments we use the same features and method as in the classification task, but compare four different entropy estimators. We again follow the recommendations of (Demšar, 2006).

**Kernel density leaf model.** For each leaf $\ell$ in the tree we use the kernel density estimate as leaf density,

**Improved Information Gain Estimates for Decision Tree Induction**
(Härdle et al., 2004)[Section 3.6],

$$p_\ell(y) = \frac{1}{n_\ell} \sum_{i=1}^{n_\ell} k_B(y - y_i^{(\ell)}),$$

where $k_B(y) = k(B^{-1}y)/\det(B)$ is the *kernel* and $B \in \mathbb{R}^{d \times d}$ is the *bandwidth matrix*. We estimate the bandwidth matrix using "Scott's rule" as $B = n_\ell^{-1/(d+4)} \hat{\Sigma}^{1/2}$, where $\hat{\Sigma} = \frac{1}{n_\ell - 1} \sum_{i=1}^{n_\ell} y_i y_i^T + \lambda I$ is the regularized *sample covariance* of all samples reaching this leaf node, and $\lambda \geq 0$ is a regularization parameter determined by model selection. As basis kernel $k : \mathbb{R}^d \to \mathbb{R}_+$ we use a multivariate standard Normal kernel. Hence $p_\ell$ is a properly normalized density.

**Setup.** We use 18 univariate and multivariate regression data sets from the UCI, StatLib, AMSTAT repositories, and one own data set, see Table 2.

The absolute continuity assumption is satisfied for most of the data sets we considered, with one notable exception: for the UCI `forestfire` data set the predictive target is one dimensional (the area of forest burned in a particular district), but is a mixed variable, being exactly zero for around half of the instances, and continuous real-valued for the other half. As such, it does not have an absolutely continuous CDF and the information gain criterion is not justified. If we would continue anyway we can have $\rho_i = 0$, leading to invalid expression of $\log 0$ in (9).

For other data sets, however, another problem occurs: *discretization* of an originally continuous variable. For example, in the classic Boston `housing` benchmark data set from the UCI repository, the predictive target (median value in 1000 USD) is discretized at a granularity of 0.1, whereas the original quantity is clearly continuous. To satisfy the absolute continuity assumption we opted to add—prior to any experiments—a sample from the uniform $U(-h/2, h/2)$ to each predictive target, where $h$ is the discretization bin width. This corresponds to a piecewise constant density in each bin, a reasonable assumption for the small bin sizes used. We do this for all data sets that have two or more identical predictive target vectors, and estimate the bin size $h$ for each dimension individually by taking the smallest positive difference between sorted values in that dimension.

As a last step, because the differential entropy is not invariant to affine transformations we standardize each output dimension of the training set individually, but undo this scaling on the target predictions before measuring the root mean squared error.

We perform model selection by first splitting the entire data set into `trainval` and `test` in proportions 60/40, and then replicate ten times the following procedure: split the `trainval` data set into `train` and `val` data sets at proportions 40/20, train a model from the `train` set and evaluate its performance on `val`. The best model parameters for each estimator is used in the final training replicates using the full `trainval` data set for training. The following parameters were fixed in all experiments: number of trees 8, minimum samples per leaf 16, number of feature tests 256, non-parametric subsampling size 256. The parameter selected was the kernel density estimation regularization $\lambda \in \{10^{-4}, 10^{-3}, 10^{-2}, 0.1, 1\}$. All models are trained on exactly the same data sets using the same parameters, except for the entropy estimator.

We estimate the expected log-likelihood $\langle LL \rangle$ for a single sample from the true distribution as average and one unit standard deviations over the ten runs, trained on `trainval` and tested on `test`. The log-likelihood is the uniform average of each individual tree model.

**Results.** The non-parametric entropy estimator yields the highest ranking holdout likelihood (average rank 1.83) compared to the other estimators. A Friedman test (Demšar, 2006) rejects the null-hypothesis of equal performance (p-value 0.0272, Iman-Davenport statistic 3.31 at $N = 18$, $k = 4$). In contrast, for the RMSE performance the null-hypothesis of equal performance is not rejected (p-value 0.350, Iman-Davenport statistic 1.12), suggesting that a better log-likelihood does not yield lower RMSE. The runtime of the 1NN estimator is a about 6-10 times larger per data set than for the Normal estimator; the entire experiments completed in 24 hours on an eight core PC.

**Discussion.** MVN-UMVUE and MVN-PLUGIN seem to perform equally well and this indicates that the misspecification error dominates the Normal entropy estimation error. Training with the 1-NN estimator yields the best log-likelihood in 10 out of 18 cases, showing that the conditional density $p(y|x)$ is captured more accurately. This does not translate into a statistically significant improvement in RMSE, where the 1-NN estimator wins 8 out of 18 cases. The lack of improvement in terms of RMSE makes sense from the point of view of empirical risk minimization.

## 4. Conclusion

We have proposed the use of recently developed entropy estimators for decision tree training. Our approach applies only to classification and regression trees using information gain as a split scoring function. While this appears limiting, this variant of decision

Improved Information Gain Estimates for Decision Tree Induction

*Table 2.* Multivariate regression results: average log-likelihood and root mean squared error (RMSE). The first set of data sets originate from the UCI repository, the second set from StatLib and AMSTAT, and the third (`kinectk2la`) is our own data set. For each data set we report the input feature dimensionality $i$, the output dimensionality $d$ and the total number of samples $N$. The average ranks (1 to 4) in terms of the holdout likelihood are also reported at the bottom of the table. For each data set we print the highest average holdout log-likelihood in boldface.

| Data set | $i/d/N$ | | Using $\hat{H}_{\text{DIAG}}$ | Using $\hat{H}_{\text{MVN-PLUGIN}}$ | Using $\hat{H}_{\text{MVN-UMVUE}}$ | Using $\hat{H}_{1\text{NN}}$ |
|---|---|---|---|---|---|---|
| PYRIM | 27/1/74 | $\langle\text{LL}\rangle$ | $-1.059 \pm 0.009$ | $-1.050 \pm 0.002$ | $-1.173 \pm 0.021$ | **-1.015±0.003** |
| | | RMSE | $0.50 \pm 0.01$ | $0.47 \pm 0.00$ | $0.61 \pm 0.04$ | $0.44 \pm 0.00$ |
| TRIAZINES | 60/1/186 | $\langle\text{LL}\rangle$ | $-1.272 \pm 0.188$ | **-1.091±0.055** | $-1.471 \pm 0.176$ | $-1.568 \pm 0.356$ |
| | | RMSE | $1.09 \pm 0.03$ | $1.07 \pm 0.03$ | $1.15 \pm 0.06$ | $1.16 \pm 0.03$ |
| MPG | 7/1/392 | $\langle\text{LL}\rangle$ | $-0.490 \pm 0.034$ | $-0.669 \pm 0.062$ | $-0.627 \pm 0.052$ | **-0.412±0.011** |
| | | RMSE | $38.97 \pm 0.55$ | $43.49 \pm 0.94$ | $44.86 \pm 0.84$ | $37.08 \pm 0.61$ |
| HOUSING | 13/1/506 | $\langle\text{LL}\rangle$ | **-0.421±0.043** | $-0.476 \pm 0.113$ | $-0.641 \pm 0.084$ | $-0.550 \pm 0.151$ |
| | | RMSE | $58.82 \pm 1.03$ | $53.85 \pm 2.42$ | $69.54 \pm 1.48$ | $61.57 \pm 1.66$ |
| MG | 6/1/1385 | $\langle\text{LL}\rangle$ | $-0.679 \pm 0.017$ | $-0.696 \pm 0.017$ | $-0.977 \pm 0.048$ | **-0.650±0.009** |
| | | RMSE | $2.91 \pm 0.03$ | $2.94 \pm 0.03$ | $3.19 \pm 0.05$ | $2.81 \pm 0.03$ |
| SPACEGA | 6/1/3107 | $\langle\text{LL}\rangle$ | $-0.928 \pm 0.026$ | $-0.978 \pm 0.005$ | $-1.147 \pm 0.005$ | **-0.861±0.017** |
| | | RMSE | $4.29 \pm 0.03$ | $4.36 \pm 0.04$ | $4.65 \pm 0.07$ | $4.31 \pm 0.04$ |
| ABALONE | 8/1/4177 | $\langle\text{LL}\rangle$ | $-1.003 \pm 0.028$ | $-0.977 \pm 0.019$ | $-1.189 \pm 0.005$ | **-0.913±0.010** |
| | | RMSE | $95.77 \pm 0.48$ | $94.95 \pm 0.69$ | $97.34 \pm 1.17$ | $94.53 \pm 0.32$ |
| CPUSMALL | 12/1/8192 | $\langle\text{LL}\rangle$ | $0.514 \pm 0.013$ | **0.526±0.031** | $0.363 \pm 0.009$ | $0.515 \pm 0.007$ |
| | | RMSE | $191.20 \pm 5.70$ | $186.12 \pm 6.43$ | $195.43 \pm 4.11$ | $267.35 \pm 15.59$ |
| CADATA | 8/1/20640 | $\langle\text{LL}\rangle$ | $-0.348 \pm 0.028$ | **-0.320±0.017** | $-0.588 \pm 0.012$ | $-0.421 \pm 0.017$ |
| | | RMSE | $10^5(50.10 \pm 0.94)$ | $10^5(47.94 \pm 0.35)$ | $10^5(52.88 \pm 0.35)$ | $10^5(53.89 \pm 0.72)$ |
| CONCRETESL | 7/3/103 | $\langle\text{LL}\rangle$ | $-3.303 \pm 0.124$ | $-4.015 \pm 0.515$ | $-3.924 \pm 0.062$ | **-3.206±0.173** |
| | | RMSE | $18.94 \pm 0.36$ | $19.00 \pm 0.51$ | $18.61 \pm 0.47$ | $17.72 \pm 0.26$ |
| SERVO | 4/1/167 | $\langle\text{LL}\rangle$ | $-0.131 \pm 0.006$ | $-0.078 \pm 0.000$ | $-0.078 \pm 0.000$ | **0.278±0.000** |
| | | RMSE | $7.20 \pm 0.00$ | $7.18 \pm 0.00$ | $7.18 \pm 0.00$ | $5.98 \pm 0.00$ |
| CONCRETESTR | 8/1/1030 | $\langle\text{LL}\rangle$ | $-0.567 \pm 0.023$ | **-0.436±0.013** | $-0.559 \pm 0.058$ | $-0.547 \pm 0.013$ |
| | | RMSE | $153.60 \pm 5.04$ | $137.01 \pm 2.53$ | $143.60 \pm 6.61$ | $150.00 \pm 4.28$ |
| PARKINSONS | 19/2/5875 | $\langle\text{LL}\rangle$ | $3.115 \pm 3.699$ | $3.522 \pm 0.201$ | $3.518 \pm 0.152$ | **3.953±0.106** |
| | | RMSE | $3.69 \pm 0.25$ | $2.30 \pm 0.04$ | $2.29 \pm 0.04$ | $7.04 \pm 0.22$ |
| FATDATA | 14/2/252 | $\langle\text{LL}\rangle$ | $-0.938 \pm 0.040$ | $-1.239 \pm 0.061$ | $-1.272 \pm 0.050$ | **-0.811±0.032** |
| | | RMSE | $12.93 \pm 0.65$ | $11.41 \pm 1.05$ | $11.31 \pm 1.02$ | $11.13 \pm 0.70$ |
| PLASMA | 12/2/315 | $\langle\text{LL}\rangle$ | $-2.722 \pm 0.078$ | $-2.891 \pm 0.072$ | $-2.876 \pm 0.133$ | **-2.540±0.098** |
| | | RMSE | $283.89 \pm 3.88$ | $287.93 \pm 4.48$ | $287.93 \pm 4.20$ | $279.36 \pm 2.54$ |
| CPS85 | 9/2/534 | $\langle\text{LL}\rangle$ | $-2.655 \pm 0.022$ | $-2.655 \pm 0.019$ | **-2.650±0.012** | $-2.684 \pm 0.029$ |
| | | RMSE | $5.64 \pm 0.07$ | $5.60 \pm 0.08$ | $5.57 \pm 0.03$ | $5.65 \pm 0.04$ |
| KUIPER | 10/2/804 | $\langle\text{LL}\rangle$ | $-1.284 \pm 0.021$ | $-1.083 \pm 0.009$ | **-0.450±0.019** | $-0.765 \pm 0.035$ |
| | | RMSE | $8907.56 \pm 25.34$ | $8924.12 \pm 14.42$ | $8923.40 \pm 9.54$ | $9161.77 \pm 38.31$ |
| KINECTK2LA | 71/9/2087 | $\langle\text{LL}\rangle$ | $-4.314 \pm 0.149$ | $-2.135 \pm 0.181$ | **-2.027±0.169** | $-2.244 \pm 0.414$ |
| | | RMSE | $0.30 \pm 0.00$ | $0.30 \pm 0.00$ | $0.30 \pm 0.00$ | $0.30 \pm 0.01$ |
| LL RANK | | | 2.72 | 2.39 | 3.06 | **1.83** |
| RMSE RANK | | | 2.56 | 2.11 | 2.89 | 2.44 |
| LL WINS | | | 1 | 4 | 3 | 10 |
| RMSE WINS | (1 TIE) | | 2 | 5 | 2 | 8 |

trees is a popular one, implemented in many software packages and used in hundreds of publications. We therefore suggest that the improvement in predictive performance derived from improved information gain estimates, while small, is useful to many. Furthermore, the changes are well-motivated and require only minor modifications to existing implementations.

**Acknowledgements.** The author thanks Jamie Shotton, Antonio Criminisi, and the anonymous reviewers for their helpful feedback.